%% file: MOSAIC_arXiv.tex
\documentclass[10pt,twocolumn,letterpaper]{article}
\usepackage[pagenumbers]{cvpr}
\usepackage{graphicx}
\usepackage{xcolor}
\usepackage{amsmath}
\usepackage{amssymb}
\usepackage{multirow}
\usepackage{etoolbox}
\usepackage{tabularx}
\usepackage{float}
\usepackage{cuted}
\usepackage{placeins}
\usepackage{needspace}
\usepackage{algorithm}
\usepackage{algpseudocode}
\usepackage{chngcntr}
\definecolor{cvprblue}{rgb}{0.21,0.49,0.74}
\usepackage[pagebackref,breaklinks,colorlinks,allcolors=cvprblue]{hyperref}
\newcommand{\suppmarker}[1]{\ifdefined\ReviewAppendixMarks\textcolor{red}{\scriptsize\textsf{[#1]}}\fi}
\newcommand{\firstparhead}[1]{\par\noindent\textbf{#1}\enspace}
\newcommand{\indentedparhead}[1]{\par\noindent\hspace*{\parindent}\textbf{#1}\enspace}
\newcommand{\openingfigure}{%
  \begin{strip}
    \centering
    \includegraphics[width=0.92\textwidth]{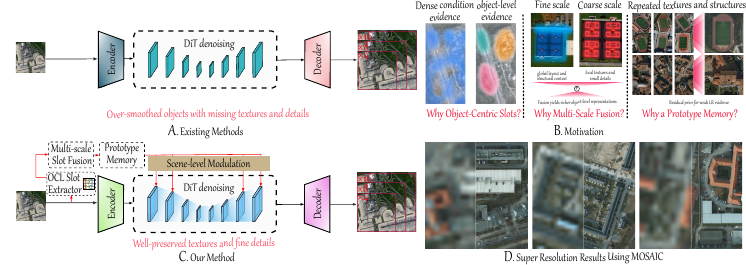}
    \captionsetup{hypcap=false}
    \captionof{figure}{MOSAIC at a glance. (a) Existing DiT-based SR primarily relies on dense LR conditioning, where salient foreground structures can be obscured by dominant backgrounds. (b) Three characteristics of remote sensing scenes motivate our design: salient foreground objects, substantial scale variation, and recurring structural patterns. (c) MOSAIC integrates object-centric slots, adaptive multi-scale fusion, and shared prototype priors to provide complementary structural guidance for generative SR. (d) Representative super-resolution results produced by MOSAIC.}
    \label{fig:opening}
  \end{strip}%
}

\def\confName{CVPR}
\def\confYear{2027}
\title{MOSAIC: Learning Object-Centric Structural Representations for Remote Sensing Image Super-Resolution}
\input{authors}

\begin{document}
\maketitle
\openingfigure

\begin{abstract}
Remote sensing image super-resolution (RSISR) requires faithfully reconstructing small foreground objects, multi-scale structures, and repetitive spatial patterns from degraded observations. Despite recent advances in generative super-resolution, existing methods often rely on dense low-resolution (LR) conditioning, which can entangle salient foreground structures with dominant backgrounds and insufficiently exploit complementary structural evidence across scales and samples. To address these limitations, we propose MOSAIC, an object-centric structural representation learning framework that complements dense spatial conditioning with compact structural guidance for generative RSISR. Specifically, MOSAIC extracts object-centric structural slots to emphasize salient foreground evidence, adaptively integrates stable cross-scale structures with fine-grained details to address substantial scale variation, and exploits recurring spatial patterns through image-guided retrieval from a shared prototype memory. The resulting representations combine image-specific evidence with shared structural priors to provide scene-level guidance for faithful reconstruction. Extensive experiments on UCMerced, AID, and RSSCN7 demonstrate consistent PSNR improvements across $\times2$ and $\times4$ settings, with gains of 2.07, 4.54, and 1.49 dB over DiT4SR at $\times4$, respectively, alongside favorable LPIPS results and improved structural preservation. The code will be publicly released.
\end{abstract}

\section{Introduction}

High-resolution remote sensing imagery supports fine-grained Earth observation, including urban monitoring, environmental analysis, and disaster assessment. Recent surveys and real-world SR studies highlight both the rapid progress of deep-learning-based super-resolution and the remaining challenges under realistic degradations \citep{moser2023hitchhiker,sun2024sdflow}. Sensor limitations, imaging conditions, and transmission constraints can obscure small objects, boundaries, and repetitive textures; recent generative SR methods likewise emphasize the difficulty of maintaining fidelity under real degradations \citep{han2026dnfsr,wu2026vosr}. Remote sensing image super-resolution (RSISR) aims to recover high-resolution observations while preserving both large-scale layouts and fine-grained structures. Remote sensing scenes exhibit large scale variation and dense spatial patterns. Small objects can be difficult to distinguish in remote sensing imagery, motivating position-guided feature extraction in object detection \citep{Wang2025PGDRFNet}. Therefore, effective RSISR requires not only plausible texture generation but also faithful reconstruction of object geometry and regular spatial arrangements.

Existing RSISR methods have evolved from convolutional and Transformer architectures to diffusion-based generative models, with recent DiT-based approaches benefiting from strong generative priors and long-range modeling \citep{rasaf,ttst,fatsnet2026}. However, they typically inject LR observations as dense spatial conditions \citep{kim2026cdm,zheng2026sdd}. As illustrated in Figure~\ref{fig:opening}(a), small objects and structural boundaries can be overwhelmed by dominant backgrounds, while recurring patterns may remain insufficiently represented under severe degradation. Consequently, strong generative priors alone do not guarantee faithful structural reconstruction. This reveals that effective RSISR depends not only on the generative capacity of the backbone, but also on how structural evidence is represented and supplied to the denoising process.

Our key insight is that effective structural conditioning requires organizing image-specific evidence while exploiting complementary information across scales and samples. As illustrated in Figure~\ref{fig:opening}(b), three observations motivate our design.
 \textbf{(i) Object-centric structural organization.} Foreground objects in remote sensing images carry critical visual and structural details for fine-grained interpretation, yet these details are easily diluted by background-dominated dense features.
 \textbf{(ii) Cross-scale complementarity.} The substantial scale variation in remote sensing scenes requires both coarse representations for global layouts and stable structures, and finer representations for local boundaries, textures, and small objects.
 \textbf{(iii) Shared structural priors.} Repetitive textures and recurring object structures across remote sensing scenes provide reusable structural priors to supplement incomplete image-specific evidence under severe degradation.

Motivated by these observations, we propose \textbf{MOSAIC}, a \textbf{M}ulti-Scale \textbf{O}bject-Centric \textbf{S}lot \textbf{A}daptation framework with \textbf{I}mage-Guided Prototype \textbf{C}onditioning for RSISR. As illustrated in Figure~\ref{fig:opening}(c), MOSAIC complements the original spatial LR condition with compact object-centric structural representations to guide the generative super-resolution process. Specifically, MOSAIC extracts object-centric structural slots from dense LR features to emphasize salient foreground evidence, integrates complementary global and local structures through adaptive multi-scale fusion, and retrieves shared structural priors to supplement incomplete image-specific evidence. The enhanced slots are then aggregated into a scene-level condition to guide structural reconstruction while preserving the pretrained generative prior. As shown in Figure~\ref{fig:opening}(d), MOSAIC recovers more coherent object structures, sharper boundaries, and finer textures.

\begin{figure*}[!t]
    \centering
    \includegraphics[width=0.88\textwidth]{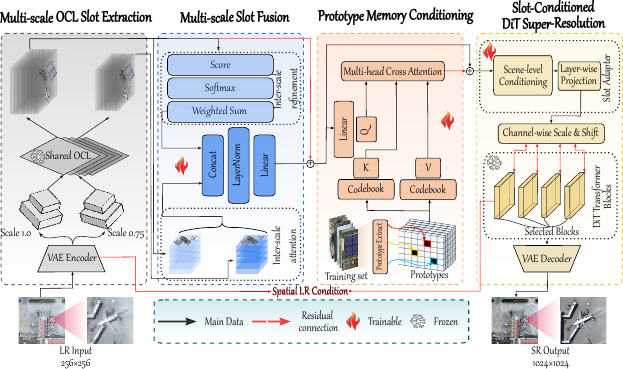}
    \caption{Overview of MOSAIC. 
    The OCL extractor organizes LR features into object-centric structural slots at multiple scales.
    MSF adaptively integrates cross-scale structural evidence with original-scale details.
    Image-guided prototype conditioning retrieves shared structural priors from an offline memory to provide residual context.
    The enhanced slots are aggregated into scene-level structural guidance, complementing the original spatial LR condition for generative super-resolution.}
    \label{fig:framework}
\end{figure*}

Our main contributions are summarized as follows:

\begin{itemize}
    \item We introduce an object-centric structural representation learning framework for generative RSISR, transforming dense LR features into compact structural slots that emphasize salient foreground objects and provide complementary structural guidance for faithful super-resolution reconstruction.

    \item We develop an adaptive multi-scale slot fusion strategy to address substantial scale variation in remote sensing scenes, selectively integrating stable cross-scale structures with original-scale detail refinement to preserve global layouts and fine-grained object details.

    \item We propose image-guided prototype conditioning that leverages recurring object structures and repetitive spatial patterns across remote sensing scenes, retrieving image-relevant structural priors from shared memory to complement incomplete observations and improve structural reconstruction.

\end{itemize}

\section{Related Work}

\subsection{Remote Sensing Image Super-Resolution}

Remote sensing image super-resolution reconstructs high-resolution images from low-resolution observations while preserving scene layouts and structures such as roads, buildings, fields, and small objects. Large-scale variation, repeated textures, and complex backgrounds make this challenging. CNN methods recover local details: SRCNN, EDSR, and RCAN use end-to-end convolution, deep residual learning, and channel attention \citep{srcnn,edsr,rcan}, while RASAF combines residual aggregation and split attention to fuse multi-level remote-sensing features \citep{rasaf}. These methods are effective for texture recovery but limited in long-range structural modeling; early CNN designs therefore struggle to maintain globally consistent layouts over large scenes.

Transformer approaches extend the receptive field: IPT transfers knowledge across restoration tasks, SwinIR and HAT use window and cross-window attention \citep{ipt,swinir,hat}, and TTST selects informative tokens to suppress irrelevant regions and model long-range ground-object relations \citep{ttst}. Although attention expands context, dense representations still mix foreground structures with redundant background content. Most approaches therefore lack compact object-level conditions, making structural evidence selection difficult in complex backgrounds. In remote sensing, narrow roads and small objects can occupy few feature tokens.

\subsection{Diffusion and DiT-based Restoration}

Diffusion models have progressed from multi-step denoising to efficient sampling and stronger architectures. DDPM formulates progressive noising and reverse denoising \citep{ddpm}, DDIM accelerates sampling while preserving quality \citep{ddim}, Latent Diffusion Models shift diffusion to a compressed latent space \citep{ldm}, and DiT processes latent tokens with Transformers for long-range spatial modeling \citep{dit}. SinSR and TSD-SR distill multi-step diffusion into one-step super-resolution \citep{sinsr,tsdsr}; these advances improve efficiency and fidelity but do not specify how structural conditions should be organized for restoration.

Iterative generative priors and global latent-space modeling support structural recovery and detail synthesis. SR3 demonstrated progressive high-frequency recovery \citep{sr3}, and DiT4SR injects low-resolution control latents into DiT for real-world super-resolution \citep{dit4sr}. However, dense single-scale conditions retain excessive background information and can weaken small objects, regular textures, and local structures; MOSAIC complements the spatial LR condition with modulation derived from multi-scale slots.

\subsection{Object-Centric Representation and Prototype Priors}

Object-centric learning compresses many input tokens into a small set of slots. MONet and IODINE learn scene decomposition and multi-object representations, while Slot Attention iteratively aggregates input features into exchangeable slots \citep{monet,iodine,slot}. Later work improves slot stability and multi-scale modeling through Smoothing Slot Attention and multi-scale object representation \citep{smoothsa,msfocl}. For super-resolution, slots can capture repeated textures or shared boundaries without instance labels.

However, image-specific slots do not explicitly summarize structures shared across samples. Prototype and codebook methods provide shared learned entries: VQ-VAE discretizes continuous features, VQ-VAE-2 expands capacity hierarchically, and VQGAN couples codebooks with adversarial synthesis \citep{vqvae,vqvae2,vqgan}. Direct quantization or feature replacement can discard image-specific details. MOSAIC therefore retrieves prototype memory as residual context for multi-scale slots before the adapter modulates selected frozen DiT blocks.

\vspace*{-0.75\baselineskip}
\section{Method}

\subsection{Overall Framework}

Figure~\ref{fig:framework} summarizes MOSAIC. The LR observation $I_{LR}$ is resized to the VAE operating resolution and encoded into a control latent $z_c$, while flow matching starts from noisy latent $z_t$. The shared OCL Slot Extractor forms compact slots at the original and resized scales; MSF fuses structural layouts with fine details. We first pretrain the OCL extractor and MSF, construct a frozen prototype memory from training slots, and then optimize the prototype cross-attention and the slot-conditioned adapter. DiT4SR also receives the spatial LR condition from the control latent. The spatial path retains the LR grid layout, and the slot path supplies scene-level channel modulation. At each sampling step, the conditioned DiT predicts the flow-matching target, the scheduler updates $z_t$, and the VAE Decoder produces $I_{SR}$.

\subsection{Object-Centric Slot Adaptation}

Remote sensing features mix multi-scale objects with complex backgrounds. Object-centric slot adaptation organizes these features into slots before forming the scene-level condition.

\indentedparhead{Step 1: Multi-scale slot extraction.}
Following the multi-scale OCL stage in Figure~\ref{fig:framework}, the shared extractor processes $z_c$ at scales $1.0$ and $0.75$ with shared parameters, producing:
\begin{equation}
S^{a}\in\mathbb{R}^{B\times K\times256},
\end{equation}
where $B$, $K$, and $256$ denote batch size, slot number, and slot dimension. The slots organize spatial evidence for local regions, textures, and layouts without masks or category labels.

\indentedparhead{Step 2: Prototype-enhanced slot conditioning.}
MSF fuses the two scales into $S_{\mathrm{msf}}$, as described in the following subsection. Prototype retrieval then produces $S_{\mathrm{pmc}}$ (Section~\ref{sec:pmc}). The adapter averages the enhanced slots and applies LayerNorm to form a scene-level condition $z$:
\begin{equation}
z=\operatorname{LayerNorm}\!\left(\frac{1}{K}\sum_{k=1}^{K}S_{\mathrm{pmc},k}\right).
\end{equation}

\begin{figure}[H]
\centering
\includegraphics[width=0.64\columnwidth]{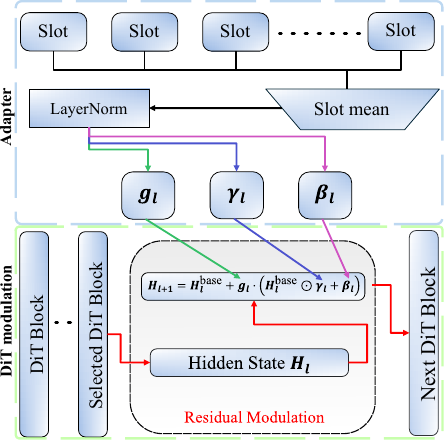}
\caption{Scene-level conditioning. The adapter summarizes the enhanced slots and predicts layer-specific channel modulation for selected frozen DiT blocks.}
\label{fig:ditocl}
\end{figure}

\indentedparhead{Step 3: Layer-wise residual modulation.}
For selected DiT block $l$, a projection produces channel-wise scale and shift terms:
\begin{equation}
[\gamma_l,\beta_l]=W_lz+b_l,
\end{equation}
and the gate $g_l$ controls residual strength, giving the gated residual update
\begin{equation}
H_{l+1}
=
H_l^{\mathrm{base}}
+
g_l
\left(
H_l^{\mathrm{base}}\odot\gamma_l+\beta_l
\right),
\label{eq:residual-modulation}
\end{equation}
where $H_l^{\mathrm{base}}$ is the base hidden state, $\odot$ denotes channel-wise multiplication, and the terms are broadcast over tokens. The gate $g_l$ is a learned scalar for block $l$, initialized to one. Zero-initialized projections make the adapter initially close to identity; residual modulation preserves the DiT token length and generation path. Pseudocode and further implementation details of the object-centric slot adaptation procedure are provided in the \textcolor{blue}{\textit{Technical Supplement}}\suppmarker{R4,R6}. Figure~\ref{fig:ditocl} illustrates the scene condition and its layer-wise residual update.

\subsection{Multi-Scale Slot Fusion}
\label{sec:msf}

\begin{figure}[H]
\centering
\includegraphics[width=0.72\columnwidth]{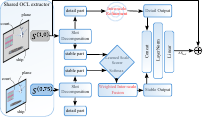}
\caption{Multi-scale slot fusion. MSF selects stable information across scales and refines base-scale details. It produces a residual update for the original slots.}
\label{fig:msf}
\end{figure}

Figure~\ref{fig:msf} summarizes Multi-Scale Slot Fusion (MSF). Coarse slots capture stable layouts and long-range relations, whereas original-scale slots retain textures, boundaries, and small objects. MSF fuses these cues in slot space and predicts a residual update for the original-scale slots.

\indentedparhead{Step 1: Slot decomposition.}
The shared OCL extractor outputs slots at both scales:
\begin{equation}
S^{1.0},S^{0.75}\in\mathbb{R}^{B\times K\times256},
\end{equation}
where shared parameters place both scales in a comparable feature space. MSF splits each $256$-D slot into branches used for cross-scale structure selection and base-scale detail refinement:
\begin{equation}
S^{a}
=
\left[
S_{\mathrm{sta}}^{a},
S_{\mathrm{det}}^{a}
\right],
\quad a\in\{1.0,0.75\},
\end{equation}
where
\begin{equation}
S_{\mathrm{sta}}^{a},S_{\mathrm{det}}^{a}
\in\mathbb{R}^{B\times K\times128},
\end{equation}
where $S_{\mathrm{sta}}^{a}$ is used for cross-scale structure selection, whereas $S_{\mathrm{det}}^{a}$ is used for current-resolution texture and boundary refinement. This architectural partition is functional rather than a claim that fixed channel subsets have inherent semantics.

\indentedparhead{Step 2: Adaptive inter-scale fusion for stable structures.}
For each slot, the Learned Scale Scorer predicts a score per scale from the stable components:
\begin{equation}
r_{k,a}
=
\mathbf{w}^{\top}S_{\mathrm{sta},k}^{a}+b,
\quad a\in\{1.0,0.75\},
\end{equation}
and softmax over scales gives:
\begin{equation}
\alpha_{k,a}
=
\frac{\exp(r_{k,a})}
{\sum_{a'\in\{1.0,0.75\}}\exp(r_{k,a'})},
\end{equation}
while the stable output is:
\begin{equation}
\hat{S}_{\mathrm{sta},k}
=
\sum_{a\in\{1.0,0.75\}}
\alpha_{k,a}S_{\mathrm{sta},k}^{a}.
\end{equation}
\par\noindent\hspace*{\parindent}Thus, each slot selects reliable structural evidence instead of fixed averaging. The weights can differ across slots within the same image.

\indentedparhead{Step 3: Intra-scale refinement for local details.}
The detail branch refines the original-scale feature $S_{\mathrm{det}}^{1.0}$ with self-attention among $K$ slots:
\begin{equation}
\hat{S}_{\mathrm{det}}
=
\operatorname{MHA}
\left(
S_{\mathrm{det}}^{1.0},
S_{\mathrm{det}}^{1.0},
S_{\mathrm{det}}^{1.0}
\right),
\end{equation}
which exchanges local context for boundaries, textures, and small objects; omitting coarse-scale details avoids over-smoothing.

\indentedparhead{Step 4: Residual fusion.}
The stable and detail outputs are concatenated, normalized, and projected:
\begin{equation}
\Delta S_{\mathrm{msf}}
=
W_o\operatorname{LayerNorm}
\left(
\left[
\hat{S}_{\mathrm{sta}},
\hat{S}_{\mathrm{det}}
\right]
\right),
\end{equation}
and is added to the original-scale slots:
\begin{equation}
S_{\mathrm{msf}}
=
S^{1.0}
+
\Delta S_{\mathrm{msf}}.
\label{eq:msf}
\end{equation}
\par\noindent\hspace*{\parindent}This residual preserves image-specific information while MSF learns complementary corrections. $S_{\mathrm{msf}}$ is passed to prototype memory conditioning; the resulting $S_{\mathrm{pmc}}$ is then passed to the adapter.

\subsection{Image-Guided Prototype Conditioning}
\label{sec:pmc}

The fused slots $S_{\mathrm{msf}}$ describe the input, but severe LR degradation can obscure objects and textures. Figure~\ref{fig:opening}(b), right, shows recurring track and rooftop patterns in training scenes. As shown in Figure~\ref{fig:framework}, PMC builds memory from fused training slots and retrieves residual context through cross-attention.

\indentedparhead{Step 1: Constructing the prototype memory.}
We flatten MSF slots from all training images and apply K-means; the $M$ centroids form the prototype memory:
\begin{equation}
P\in\mathbb{R}^{M\times256}.
\end{equation}
\par\noindent\hspace*{\parindent}The centroids summarize recurring structures and spatial patterns in the training distribution as shared residual priors.

\indentedparhead{Step 2: Retrieving shared structural context.}
The memory $P$ remains frozen. For the current image, $S_{\mathrm{msf}}$ provides queries and $P$ keys and values:
\begin{gather}
H_p=\operatorname{PMC}(Q=S_{\mathrm{msf}},K=P,V=P),\\
S_{\mathrm{pmc}}=S_{\mathrm{msf}}+\alpha H_p,
\end{gather}
where PMC retrieves context relevant to the current slots. Each slot retrieves its own weighted combination of centroids from the frozen memory.

\indentedparhead{Step 3: Residual conditioning.}
With $\alpha=0.10$, the normalized memory output is added residually to $S_{\mathrm{msf}}$, complementing image-specific evidence without replacing it. Only the PMC projections and adapter are updated; $S_{\mathrm{pmc}}$ then conditions frozen DiT blocks through the adapter. The memory is constructed offline from training slots, remains frozen thereafter, and is used at both adaptation and inference.

\section{Experiments}

\subsection{Experimental Setup}

\firstparhead{Datasets.} We evaluate MOSAIC on UCMerced \citep{ucmerced}, AID \citep{aid}, and RSSCN7 \citep{rsscn7} at $\times2$ and $\times4$, reporting PSNR, SSIM, LPIPS \citep{lpips}, MANIQA \citep{maniqa}, and CLIP-IQA \citep{clipiqa}. The no-reference MANIQA and CLIP-IQA metrics are evaluated only at $\times4$. Methods are compared on matched LR and HR test pairs. PSNR and SSIM are computed in RGB after cropping a four-pixel border.

\indentedparhead{Implementation Details.}
Data are partitioned before LR generation, so held-out images are excluded from both optimization and the construction of the frozen prototype memory. Main comparisons and the component ablation use test splits. The remaining ablations use validation splits. Detailed data splits, LR generation, and evaluation protocols are provided in the \textcolor{blue}{\textit{Technical Supplement}}\suppmarker{R1}. Implemented in PyTorch on 4 Tesla V100-SXM2-32GB GPUs, MOSAIC uses AdamW, FP16 mixed precision, batch size 1/GPU, and a final-stage learning rate of $5\times10^{-7}$ on all three datasets. Unless noted, we use 8 256-D slots (3 iterations), MSF scales $[1.0,0.75]$, 16 prototypes, 4-head prototype cross-attention (64-D heads), residual scale 0.1, adapter placement at blocks $[2,8,12,16]$, and 10 sampling steps.

\subsection{Main Results}

\firstparhead{Quantitative results.} Table~\ref{tab:psnr-ssim} reports the full-reference results (PSNR and SSIM). MOSAIC achieves the best PSNR among the compared methods at both scales. Under the $\times4$ setting, it improves PSNR over DiT4SR by 2.07 dB on UCMerced, 4.54 dB on AID, and 1.49 dB on RSSCN7; the corresponding $\times2$ gains are 2.90 dB, 5.40 dB, and 2.25 dB. The ablations below examine how the individual components contribute to these gains.

Table~\ref{tab:perceptual} reports the perceptual-quality results (LPIPS, MANIQA, and CLIP-IQA). MOSAIC obtains the lowest LPIPS on both RSSCN7 and AID. The MANIQA and CLIP-IQA results vary across datasets while remaining competitive. Together, these comparisons indicate strong perceptual similarity alongside competitive no-reference image quality. LPIPS uses the HR reference, while MANIQA and CLIP-IQA evaluate only the reconstruction. The resulting rankings can differ.

\begin{table}[!t]
\centering
\footnotesize
\setlength{\tabcolsep}{2pt}
\renewcommand{\arraystretch}{1.0}

\caption{PSNR and SSIM on three test splits under $\times2$ and $\times4$. \textcolor{red}{$\blacksquare$} best; \textcolor{blue}{$\blacksquare$} second best.}
\label{tab:psnr-ssim}
\begin{tabular}{@{}lrrrr@{}}
        \toprule
        Method & \multicolumn{2}{c}{$\times2$}
        & \multicolumn{2}{c}{$\times4$}\\
        \cmidrule(lr){2-3}
        \cmidrule(lr){4-5}
        & PSNR $\uparrow$ & SSIM $\uparrow$
        & PSNR $\uparrow$ & SSIM $\uparrow$\\
        \midrule

        \multicolumn{5}{@{}l}{\textit{UCMerced}}\\
        Real-ESRGAN & \textcolor{blue}{\underline{\textbf{26.8314}}} & \textcolor{blue}{\underline{\textbf{0.8023}}} & \textcolor{blue}{\underline{\textbf{23.3802}}} & 0.6215\\
        KDSR-GANV2  & 26.2297 & 0.7842 & 23.1964 & 0.6113\\
        LDL         & 25.7230 & 0.8004 & 22.9224 & \textcolor{red}{\textbf{0.6339}}\\
        MM-RealSR   & 24.5995 & 0.7895 & 22.1259 & \textcolor{blue}{\underline{\textbf{0.6255}}}\\
        DiT4SR      & 26.4208 & 0.7922 & 22.1419 & 0.5579\\
        \textbf{MOSAIC}
                      & \textcolor{red}{\textbf{29.3194}} & \textcolor{red}{\textbf{0.8473}}
                      & \textcolor{red}{\textbf{24.2119}} & 0.6100\\
        \midrule

        \multicolumn{5}{@{}l}{\textit{AID}}\\
        Real-ESRGAN & \textcolor{blue}{\underline{\textbf{27.6707}}} & \textcolor{blue}{\underline{\textbf{0.8104}}} & \textcolor{blue}{\underline{\textbf{23.8303}}} & 0.6153\\
        KDSR-GANV2  & 27.0080 & 0.7902 & 23.6919 & 0.6097\\
        LDL         & 26.1695 & 0.8063 & 23.2775 & \textcolor{blue}{\underline{\textbf{0.6273}}}\\
        MM-RealSR   & 25.6580 & 0.8031 & 22.6745 & 0.6200\\
        DiT4SR      & 27.0105 & 0.8016 & 22.6609 & 0.5534\\
        \textbf{MOSAIC}
                      & \textcolor{red}{\textbf{32.4056}} & \textcolor{red}{\textbf{0.8910}}
                      & \textcolor{red}{\textbf{27.1982}} & \textcolor{red}{\textbf{0.6908}}\\
        \midrule

        \multicolumn{5}{@{}l}{\textit{RSSCN7}}\\
        Real-ESRGAN & \textcolor{blue}{\underline{\textbf{27.7757}}} & 0.7801 & 24.8375 & 0.6018\\
        KDSR-GANV2  & 27.6795 & 0.7703 & \textcolor{blue}{\underline{\textbf{24.8750}}} & 0.5997\\
        LDL         & 26.1831 & 0.7779 & 24.2620 & \textcolor{red}{\textbf{0.6089}}\\
        MM-RealSR   & 26.1496 & \textcolor{blue}{\underline{\textbf{0.7827}}} & 23.7773 & \textcolor{blue}{\underline{\textbf{0.6073}}}\\
        DiT4SR      & 27.3353 & 0.7740 & 23.8106 & 0.5430\\
        \textbf{MOSAIC}
                      & \textcolor{red}{\textbf{29.5817}} & \textcolor{red}{\textbf{0.8249}}
                      & \textcolor{red}{\textbf{25.2957}} & 0.5875\\
        \bottomrule
    \end{tabular}
\end{table}

\begin{table}[!t]
\centering
\footnotesize
\setlength{\tabcolsep}{2pt}
\renewcommand{\arraystretch}{1.0}

\caption{Perceptual quality on RSSCN7 and AID test splits at $\times4$. \textcolor{red}{$\blacksquare$} best; \textcolor{blue}{$\blacksquare$} second best.}
\label{tab:perceptual}
\begin{tabular}{@{}lrrr@{}}
        \toprule
        Method & LPIPS $\downarrow$
        & MANIQA $\uparrow$
        & CLIP-IQA $\uparrow$\\
        \midrule

        \multicolumn{4}{@{}l}{\textit{RSSCN7}}\\
        Real-ESRGAN & \textcolor{blue}{\underline{\textbf{0.3199}}} & \textcolor{blue}{\underline{\textbf{0.3276}}} & 0.4446\\
        KDSR-GANV2  & 0.3554 & 0.3088 & 0.4336\\
        LDL         & 0.3231 & 0.2911 & 0.4579\\
        DiT4SR      & 0.3426 & 0.3260 & \textcolor{red}{\textbf{0.5391}}\\
        \textbf{MOSAIC}
                       & \textcolor{red}{\textbf{0.2737}} & \textcolor{red}{\textbf{0.3297}} & \textcolor{blue}{\underline{\textbf{0.4674}}}\\
        \midrule

        \multicolumn{4}{@{}l}{\textit{AID}}\\
        Real-ESRGAN & 0.2803 & \textcolor{red}{\textbf{0.3243}} & 0.4663\\
        KDSR-GANV2  & 0.3011 & 0.3154 & 0.4688\\
        LDL         & 0.2847 & 0.2801 & 0.4679\\
        DiT4SR      & \textcolor{blue}{\underline{\textbf{0.2739}}} & \textcolor{blue}{\underline{\textbf{0.3207}}} & \textcolor{red}{\textbf{0.5653}}\\
        \textbf{MOSAIC}
                       & \textcolor{red}{\textbf{0.2149}} & 0.2849 & \textcolor{blue}{\underline{\textbf{0.4862}}}\\
        \bottomrule
    \end{tabular}
\end{table}

\begin{figure*}[!t]
    \centering
    \includegraphics[width=\textwidth]{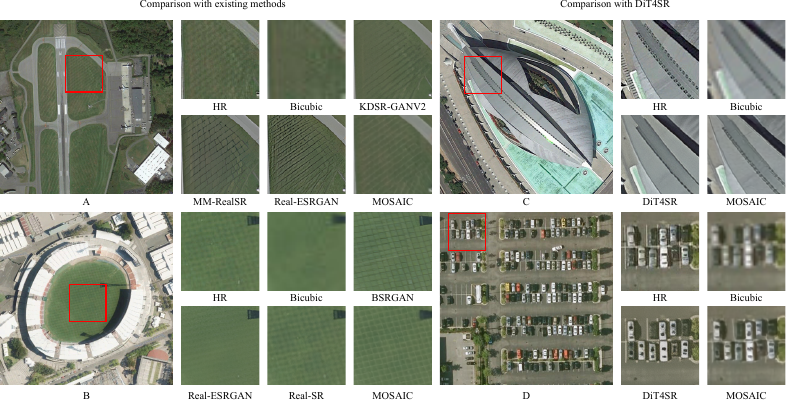}
    \caption{Qualitative super-resolution comparisons. Left: MOSAIC is compared with Bicubic and existing SR methods on Airport and Stadium scenes. Right: MOSAIC is compared directly with DiT4SR on Center and Parking scenes. Red boxes mark the enlarged regions.}
    \label{fig:qualitative}
\end{figure*}

\indentedparhead{Qualitative results.}
Figure~\ref{fig:qualitative} compares reconstructions on Airport, Stadium, Center, and Parking. In Airport (A), the faint diagonal stripes in the HR grass are blurred by bicubic interpolation, while some SR methods render them as prominent dark bands. MOSAIC recovers the stripe direction with more subdued contrast. In Stadium (B), \mbox{MOSAIC} brings out the repeated grid across the pitch while keeping its contrast closer to the HR crop than the prominent grid produced by BSRGAN.

In Center (C), the HR crop contains a row of closely spaced dark marks running diagonally across the roof. DiT4SR smooths much of this row into a thin, nearly continuous line. MOSAIC retains a visible sequence of separate marks along the diagonal, together with faint variations on the adjacent roof surface. The broader roof edges are visible in both reconstructions. The difference is in the smaller, repeated detail along those edges.

{\clubpenalty=10000 In Parking (D), DiT4SR reconstructs several vehicles as similar rectangular shapes with strong, rounded dark patches. MOSAIC retains differences in the outlines and in the size and placement of dark regions along neighboring vehicles. For example, the dark areas on the white vehicle near the center of the lower row are more elongated in MOSAIC, closer to the variation along its body in the HR crop. The roof and parking examples illustrate two kinds of detail: the repetition of small marks on one large surface and the variation between adjacent small objects. Additional visual comparisons with DiT4SR and per-image metrics are provided in Sec.~A of the \textcolor{blue}{\textit{Technical Supplement}}.\par}

\subsection{Ablation Studies}

\indentedparhead{Main-component ablation.}
Table~\ref{tab:main-ablation} shows that the OCL-conditioned adapter improves PSNR over DiT4SR by 1.6790~dB. The configuration with MSF gains a further 0.2820~dB and improves SSIM and LPIPS. The configuration with PMC yields a further PSNR gain of 0.109~dB, increases SSIM by 0.0073, and reduces LPIPS by 0.0073.

\begin{table}[H]
\centering
\footnotesize
\setlength{\tabcolsep}{2pt}
\renewcommand{\arraystretch}{1.2}

\caption{Main-component ablation (UCMerced $\times4$ test). The first row is DiT4SR. OCL conditions the adapter in the remaining rows.}
\label{tab:main-ablation}
\begin{tabular*}{\linewidth}{@{\extracolsep{\fill}}ccc rrr@{}}
    \hline
    OCL & MSF & PMC & PSNR $\uparrow$ & SSIM $\uparrow$ & LPIPS $\downarrow$ \\
    \hline
    & & & 22.1419 & 0.5579 & 0.2940 \\
    \checkmark & & & 23.8209 & 0.5943 & 0.2417 \\
    \checkmark & \checkmark & & 24.1029 & 0.6027 & 0.2315 \\
    \checkmark & \checkmark & \checkmark & \textbf{24.2119} & \textbf{0.6100} & \textbf{0.2242} \\
    \hline
\end{tabular*}
\end{table}

\indentedparhead{MSF-variant ablation.}
Table~\ref{tab:msf-ablation} compares different ways to combine multi-scale slots. Direct averaging provides only a small improvement over the single-scale baseline. Inter-scale fusion gives a larger gain, demonstrating the benefit of learned cross-scale interaction over fixed averaging. Full MSF reaches 24.2510 dB PSNR, 0.6386 SSIM, and 0.2210 LPIPS. It improves all three metrics over inter-scale-only fusion, supporting the addition of original-scale detail refinement. Relative to single-scale conditioning, full MSF increases PSNR by 0.0990~dB and SSIM by 0.0112, and decreases LPIPS by 0.0115.

\begin{table}[H]
\centering
\footnotesize
\setlength{\tabcolsep}{2pt}
\renewcommand{\arraystretch}{1.2}

\caption{MSF-variant ablation (UCMerced $\times4$ validation).}
\label{tab:msf-ablation}
\begin{tabular*}{\linewidth}{@{\extracolsep{\fill}}l rrr@{}}
    \hline
    MSF variant & PSNR $\uparrow$ & SSIM $\uparrow$ & LPIPS $\downarrow$ \\
    \hline
    Single-scale      & 24.1520 & 0.6274 & 0.2325 \\
    Multi-scale Avg   & 24.1620 & 0.6282 & 0.2323 \\
    Inter-scale Only  & 24.2440 & 0.6368 & 0.2215 \\
    Full MSF          & \textbf{24.2510} & \textbf{0.6386} & \textbf{0.2210} \\
    \hline
\end{tabular*}
\end{table}

\Needspace{3\baselineskip}
\indentedparhead{Effectiveness of OCL slot information.}
In Table~\ref{tab:ocl-slot-input}, we keep the checkpoint and LR images fixed and vary only the slots fed to the adapter. Matched slots improve PSNR by 1.4273~dB over shuffled slots and 1.5188~dB over zero slots, with higher SSIM and lower LPIPS in both cases. Shuffling supplies the same number of learned slots from other images, breaking their correspondence to the current input.

\begin{table}[H]
\centering
\footnotesize
\setlength{\tabcolsep}{2pt}
\renewcommand{\arraystretch}{1.2}
\caption{Single-scale OCL slot input comparison (UCMerced $\times4$ validation).}
\label{tab:ocl-slot-input}
\begin{tabular*}{\columnwidth}{@{\extracolsep{\fill}}lrrr@{}}
\toprule
OCL slot input & PSNR $\uparrow$ & SSIM $\uparrow$ & LPIPS $\downarrow$ \\
\midrule
Matched to LR image & \textbf{24.0300} & \textbf{0.6244} & \textbf{0.2400} \\
Shuffled across images & 22.6027 & 0.5828 & 0.2844 \\
All-zero slots & 22.5112 & 0.5795 & 0.2857 \\
\bottomrule
\end{tabular*}
\end{table}

The drop under shuffling shows that the scene-level condition carries image-specific slot information that the adapter uses in reconstruction.

\begin{figure}[H]
\centering
\includegraphics[width=\columnwidth,height=102.44687pt]{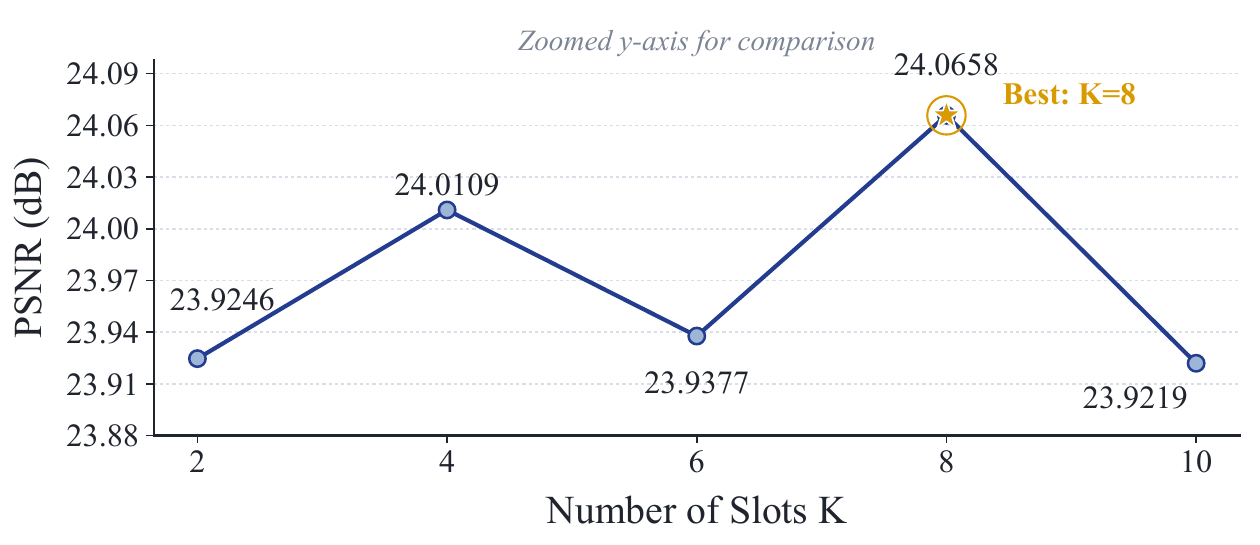}
\caption{Slot-number ablation (UCMerced $\times4$ validation).}
\label{fig:slot-number-ablation}
\end{figure}

\indentedparhead{Slot-number ablation.}
Figure~\ref{fig:slot-number-ablation} evaluates the number of object-centric slots. PSNR varies non-monotonically across the tested values and reaches its maximum of 24.0658 dB at $K=8$; it decreases at $K=10$. This result motivates our choice of $K=8$ for the final model.

\begin{figure}[H]
\centering
\includegraphics[width=\columnwidth,height=104.54492pt]{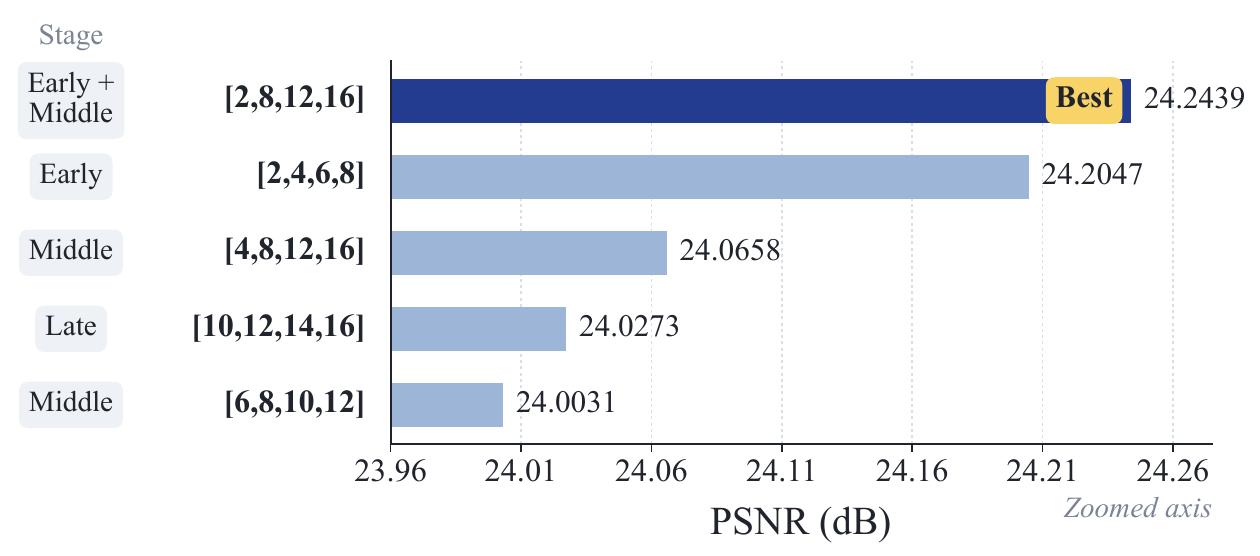}
\caption{Adapter-layer ablation (UCMerced $\times4$ validation).}
\label{fig:adapter-layers-ablation}
\end{figure}

\indentedparhead{Adapter-placement ablation.}
Figure~\ref{fig:adapter-layers-ablation} evaluates different adapter placements. The best result is obtained when the adapter modulates early and middle DiT blocks. This configuration, using blocks $[2,8,12,16]$, achieves the best PSNR of 24.2439 dB. Early blocks receive the structural condition before the denoising features become highly specialized. Middle blocks further refine the recovered content. This placement provides a better balance between structural control and texture reconstruction than using only a limited part of the denoising network.

\begin{figure}[H]
\centering
\includegraphics[width=\columnwidth,height=92.6179pt]{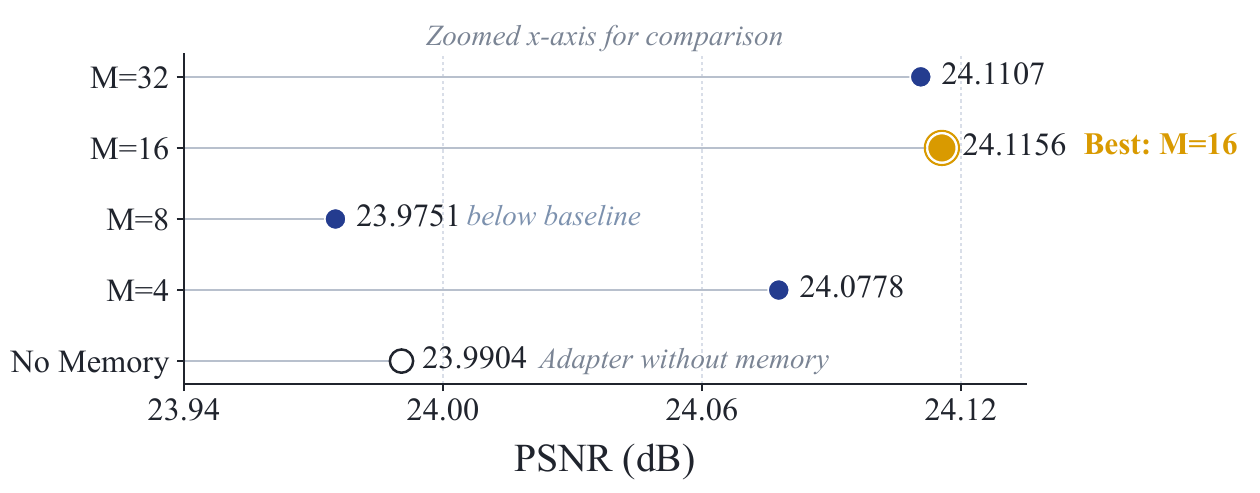}
\caption{Prototype-memory-size ablation (UCMerced $\times4$ validation).}
\label{fig:memory-ablation}
\end{figure}

\indentedparhead{Prototype-memory-size ablation.}
Figure~\ref{fig:memory-ablation} evaluates the number of prototypes in the memory. Performance reaches its peak at $M=16$, with a PSNR of 24.1156 dB. Other memory sizes yield lower performance under this ablation protocol; we therefore use 16 prototypes in the final model.

\subsection{Slot Attention Visualization}

\begin{figure}[H]
    \centering
    \includegraphics[width=\columnwidth]{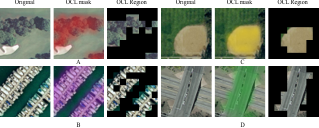}
    \caption{Slot attention responses. Each example shows an input, its learned slot response, and a localized response; slots attend to complementary structural regions.}
    \label{fig:slots}
\end{figure}

Each example in Figure~\ref{fig:slots} contains a real-world input, its learned slot response, and the corresponding localized response. Different slots attend to complementary structural regions. These responses illustrate how the OCL extractor represents salient structures in distinct slots. These masks show extractor responses before scene-level aggregation.

\vspace{0pt}

\subsection{Structural Generalization Beyond Remote Sensing}

\begin{figure}[H]
\centering
\includegraphics[width=\columnwidth]{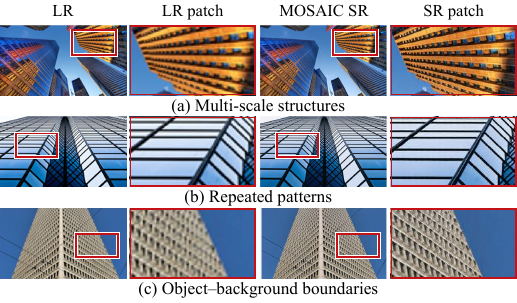}
\caption{Urban100 comparisons. Red boxes mark the enlarged regions.}
\label{fig:urban100-structure}
\end{figure}

Remote sensing and natural images share structures at multiple scales, repetitive patterns, and object boundaries. To examine how MOSAIC reconstructs these structures in natural images, we select three building scenes from Urban100~\citep{urban100}. Figure~\ref{fig:urban100-structure}(a) shows both the overall building outline and facade details. Rows of dark openings are easier to distinguish in the reconstructed crop. In (b), the repetitive horizontal window frames and their intersections with the slanted frame are clearer. In (c), windows near the building--sky boundary are easier to distinguish. These examples show how MOSAIC recovers local structures in natural images of buildings. More examples are provided in Sec.~B of the \textcolor{blue}{\textit{Technical Supplement}}.

\section{Conclusion}

In this paper, we presented MOSAIC, an object-centric structural representation learning framework for generative remote sensing image super-resolution. By organizing salient foreground evidence into structural slots and complementing it with cross-scale information and shared structural priors, MOSAIC provides structured guidance for faithful reconstruction. Experiments on three remote sensing datasets show consistent PSNR gains at $\times2$ and $\times4$, with the lowest LPIPS among the compared methods on RSSCN7 and AID at $\times4$. Qualitative analyses further illustrate improved structural preservation and potential transferability beyond remote sensing.

\vspace{-0.25\baselineskip}
\textbf{Limitations and Future Work.} The adapter averages the enhanced slots into a single scene-level vector, so all spatial tokens within each selected DiT block receive the same modulation. This limits the adapter's ability to use different slot cues in different regions. Future work could let spatial tokens attend directly to individual slots and evaluate whether this improves local structure recovery in remote sensing and natural images.

\newcommand{\mosaicLLMUsageStatement}{%
    \section*{LLM Usage Statement}
    We used OpenAI ChatGPT to assist with editing and polishing author-written text, refining the related-work citations, and editing author-written code. The authors take full responsibility for all submitted content.\par
}

\mosaicLLMUsageStatement

{
    \small
    \bibliographystyle{ieeenat_fullname}
    \bibliography{mosaic_references}
}
\end{document}

%% file: authors.tex
\author{%
\normalfont
\begin{tabular}{c}
\textbf{Liuqian Wang\textsuperscript{1}, Hanyu Xie\textsuperscript{1,\textdagger}, Rongzhen Zhao\textsuperscript{2,\textdaggerdbl}, Qi Ming\textsuperscript{3}, Chunjie Ma\textsuperscript{4},}\\
\textbf{Wentao Chao\textsuperscript{1}, Yufei Gao\textsuperscript{1}, Chengming Liu\textsuperscript{1}, Lei Shi\textsuperscript{1}}\\[0.6em]
\textsuperscript{1}Zhengzhou University, Henan, China\\
\textsuperscript{2}Aalto University, Finland\\
\textsuperscript{3}Beijing University of Technology, Beijing, China\\
\textsuperscript{4}Qilu University of Technology, Shandong, China\\[0.6em]
\textsuperscript{\textdagger}Project lead, \textsuperscript{\textdaggerdbl}Corresponding Author
\end{tabular}%
}